\documentclass[10pt,twocolumn,letterpaper]{article}

\usepackage[pagenumbers]{iccv} 

\definecolor{iccvblue}{rgb}{0.21,0.49,0.74}
\usepackage[pagebackref,breaklinks,colorlinks,allcolors=iccvblue]{hyperref}
\usepackage{graphicx}
\usepackage{amsmath}
\usepackage{amsthm}
\usepackage{makecell}
\usepackage{longtable}
\usepackage{multirow}
\usepackage{multicol}
\usepackage{lineno}
\usepackage{booktabs}
\usepackage{caption}
\usepackage{epsfig}
\usepackage{amssymb}
\usepackage{epsfig}
\usepackage{afterpage}
\usepackage{cuted}
\usepackage{float}
\usepackage{bbding}
\usepackage{bm}
\usepackage{setspace}
\def\paperID{} 
\def\confName{ICCV}
\def\confYear{2025}

\title{Aligning Anime Video Generation with Human Feedback}

\author{
    Bingwen Zhu$^{1,2}$\qquad
    Yudong Jiang$^{2}$\qquad
    Baohan Xu$^{2}$\qquad
    Siqian Yang$^{2}$\qquad 
    Mingyu Yin$^{2}$\qquad \\
    Yidi Wu$^{1,2}$\qquad
    Huyang Sun$^{2}$\qquad
    Zuxuan Wu$^{1^{\dagger}}$\qquad
    \\
    $^{1}$Fudan University\qquad
    $^{2}$Bilibili Inc.
}

\begin{document}
\maketitle
\begin{abstract}
Anime video generation faces significant challenges due to the scarcity of anime data and unusual motion patterns, leading to issues such as motion distortion and flickering artifacts, which result in misalignment with human preferences. Existing reward models, designed primarily for real-world videos, fail to capture the unique appearance and consistency requirements of anime. In this work, we propose a pipeline to enhance anime video generation by leveraging human feedback for better alignment. Specifically, we construct the first multi-dimensional reward dataset for anime videos, comprising 30k human-annotated samples that incorporating human preferences for both visual appearance and visual consistency. Based on this, we develop AnimeReward, a powerful reward model that employs specialized vision-language models for different evaluation dimensions to guide preference alignment. Furthermore, we introduce Gap-Aware Preference Optimization (GAPO), a novel training method that explicitly incorporates preference gaps into the optimization process, enhancing alignment performance and efficiency. Extensive experiment results show that AnimeReward outperforms existing reward models, and the inclusion of GAPO leads to superior alignment in both quantitative benchmarks and human evaluations, demonstrating the effectiveness of our pipeline in enhancing anime video quality. Our code and dataset are publicly available at \url{https://github.com/bilibili/Index-anisora}.
\end{abstract}    
\section{Introduction}
\label{sec:intro}
\begin{figure*}[!ht]
  \centering
  \includegraphics[width=\linewidth]{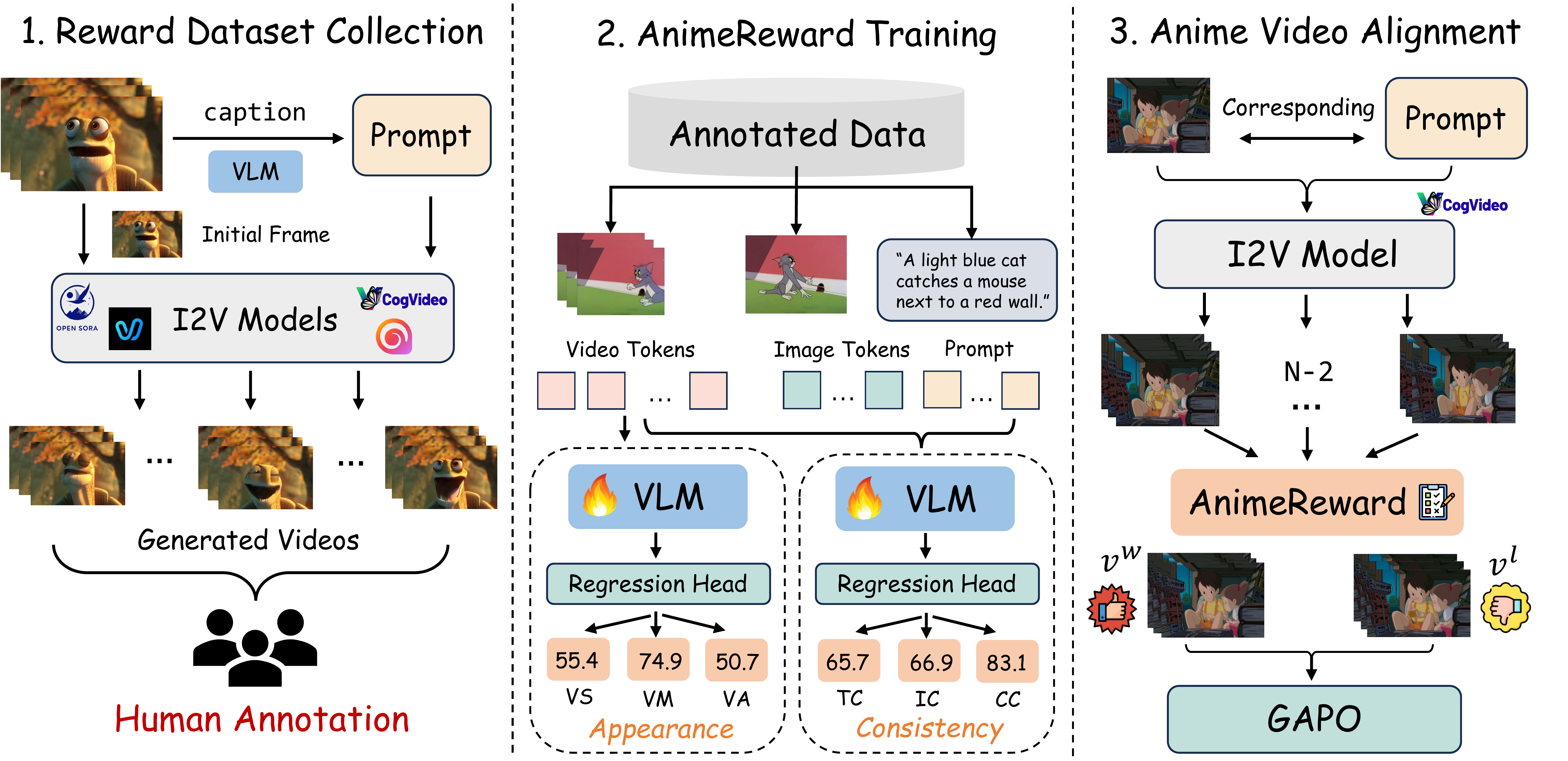}
  \caption{\textbf{Overview of our alignment pipeline.} (1) Reward Dataset Collection. We generate anime videos using 5 advanced I2V models conditioned on an initial frame and prompt. Human annotators then evaluate the generated videos across multiple dimensions. (2) AnimeReward Training. The annotated data is used to train a specialized reward model - AnimeReward. We employ dimension-specific models to separately regress preference scores for visual appearance (VS: visual smoothness, VM: visual dynamics, VA: visual appeal) and visual consistency (TC: text-video consistency, IC: image-video consistency, CC: character consistency). (3) Anime Video Alignment: Given a prompt and a corresponding anime image, the baseline I2V model generates multiple video outputs. AnimeReward then scores these videos based on preference. The best and worst videos are selected to form preference pairs for GAPO alignment training. }
  \label{fig:overview}
\end{figure*}
With the recent emergence of various powerful video generation models~\cite{yang2024cogvideox,wang2025wan,kong2024hunyuanvideo,zheng2024open,lin2024open,vidu,minimax,polyak2024movie,ma2025stepvideot2vtechnicalreportpractice,jiang2024exploring}, both industry and academia are eagerly anticipating the arrival of an era where one-click video creation becomes a reality. Anime, as a significant form of visual expression in video media, plays an indispensable role in human daily life and entertainment.  
However, due to the scarcity of anime data and differences in motion patterns, current video generation models often exhibit significant motion distortions and flickering artifacts when applied to anime video generation, resulting in the generation quality falls far short of human expectations. Generating anime videos solely from prompts is challenging because video models are pretrained on significantly more real-world data than anime data, leading to outputs that favor realism. To overcome this, we employ the Image-to-Video (I2V) framework~\cite{yang2024cogvideox,jiang2024exploring}, specifying the first frame to constrain the initial scene and style of generated videos.

Reinforcement Learning from Human Feedback (RLHF) is initially proposed as a paradigm in Natural Language Processing (NLP), demonstrating remarkable alignment capabilities in Large Language Models (LLMs). Recent advancements~\cite{fan2023dpok,black2023training,prabhudesai2023aligning,xu2023imagereward,wallace2024diffusion,yang2024using,yuan2025self} in image generation proved that RLHF remains effective in aligning generative models with human preferences. Building on this, several studies have been proposed to align video generation models with human preferences. Early approaches~\cite{yuan2024instructvideo,li2024t2v} relied on existing image-based reward models~\cite{wu2023human,kirstain2023pick} for feedback signals; however, these methods failed to account for key temporal factors in videos, such as motion dynamics and temporal coherence. To address this limitation, subsequent researches~\cite{zhang2024onlinevpo,liu2024videodpo} adopted video evaluation systems~\cite{huang2024vbench,he2024videoscore}, using the aggregated scores as proxies for human preferences. However, these evaluation metrics are typically objective and lack sufficient discriminatory power to effectively differentiate good and bad videos, making them inadequate for capturing subjective preferences. More recent studies~\cite{wang2024lift,xu2024visionreward,liu2025improving} have collected large-scale video preference datasets and trained Vision-Language Models (VLMs) as video reward models. Nevertheless, these models are primarily designed for Text-to-Video (T2V) tasks and, due to domain discrepancies, struggle to accurately assess anime videos, leading to evaluations that diverge from actual human preferences.

To address these challenges, we propose a specialized anime video alignment pipeline, as shown in Fig.~\ref{fig:overview}. First, we collect and curate the first higt-quality reward dataset in the anime domain comprising 30k human-annotated anime videos. The annotations are categorized into two primary aspects: \textbf{\textit{Visual Appearance}} and \textbf{\textit{Visual Consistency}}. The visual appearance evaluation encompasses key dimensions of visual smoothness, visual motion, and visual appeal, only focusing on video frames. In contrast, the visual consistency assessment extends beyond text-video consistency by incorporating image-video consistency, which is crucial for the Image-to-Video (I2V) task, and character consistency, which is unique to anime videos. By integrating these 6 dimensions across both aspects, the reward dataset delivers a comprehensive evaluation of anime video quality, ensuring more accurate human preference for the reward training. Based on the annotated data, we develop \textbf{AnimeReward}, a reliable multi-dimensional reward system specifically designed for aligning anime video generation. Previous approaches have used a unified vision-language model to regress preferences for all dimensions. However, we argue that the characteristics of a video that deserve attention vary significantly across different dimensions. Therefore, we employ specialized vision-language models for each dimension to better regress human preferences. 
Our video alignment training is based on Direct Preference Optimization (DPO)~\cite{rafailov2023direct}, which enables the model to learn human preferences from reward signals derived from positive and negative samples in preference pairs.
With AnimeReward, we can automatically obtain preference feedback for any anime video, facilitating the creation of preference sample pairs for alignment. Drawing inspiration from the intuition that preference pairs with larger gaps between samples should exert a greater impact on the training, we introduce a novel method called \textbf{Gap-Aware Preference Optimization (GAPO)}. This approach explicitly integrates the preference gap into the loss function to improve the alignment efficiency and performance.

In summary, our contributions are as follows:
\begin{itemize}
    \item \textbf{First Reward Dataset for Anime Videos}: We construct the first reward dataset for anime video, evaluating visual appearance and consistency to provide a comprehensive framework for assessing anime video quality.
    \item \textbf{Multi-Dimensional Anime Reward Model}: We develop AnimeReward, a reward model specifically designed for anime videos, employing specialized models for different dimensions to ensure more accurate preference alignment with human evaluations.
    \item \textbf{High quality Anime Video Alignment}: We introduce Gap-Aware Preference Optimization (GAPO), which outperform other approaches across multiple benchmarks and achieve the best results in human evaluations. Experiments demonstrate that utilizing only data generated by the baseline I2V model, our alignment pipeline significantly enhances the quality of anime generation, achieving better alignment with human preferences. 
\end{itemize}

\section{Related Works}
\label{sec:related}
\noindent{\textbf{Evaluation for Video Generation.}}
Evaluating the quality of generated videos and assigning reward scores is crucial for aligning generative models with human preferences. 
Early methods relied on metrics such as FVD~\cite{unterthiner2018towards} and CLIP~\cite{radford2021learning} to assess the realism and semantic consistency of generated videos. However, these metrics are far from sufficient to comprehensively judge video quality. Recently, VBench~\cite{huang2024vbench} introduced a more holistic evaluation framework, assessing videos across 16 dimensions covering both quality and semantic alignment. EvalCrafter~\cite{liu2024evalcrafter} proposed a benchmark with 700 prompts and employed human ratings for more thorough evaluation.
In the image generation domain, some studies~\cite{xu2023imagereward,kirstain2023pick,wu2023human,liang2024rich} have demonstrated that training CLIP-based models on human preference datasets can improve alignment between evaluation scores and human judgments. Inspired by this, works such as T2VQA~\cite{kou2024subjective} and VideoScore~\cite{he2024videoscore} leveraged human-annotated video ratings to train video quality assessment models. With the advancement of large Vision-Language Models (VLMs)~\cite{achiam2023gpt,lin2024vila,team2024gemini,hong2024cogvlm2,wang2024qwen2}, recent researches~\cite{wang2024lift,xu2024visionreward,liu2025improving} employed these models to perform score regression or preference learning across multiple dimensions, simulating human evaluation of videos. 
Nevertheless, these efforts primarily focus on open-domain video generation and lack dedicated evaluation dimensions for the I2V task in the anime domain. To bridge this gap, we construct a human preference anime dataset for the training of anime reward models. Furthermore, we utilize multiple VLMs to perform preference regression on individual dimensions, enabling accurate and highly human-aligned assessments of anime video quality.

\noindent{\textbf{RLHF in Generative Models.}} 
Reinforcement Learning from Human Feedback (RLHF) has emerged as a widely adopted strategy for aligning generative models with human preferences, leading substantial enhancements in visual generation quality. In the field of image generation, researchers have explored various RLHF-driven techniques, primarily including gradient-based optimization using reward models~\cite{clark2023directly,prabhudesai2023aligning,xu2023imagereward}, direct optimization on human preference data~\cite{wallace2024diffusion,yang2024using,liang2024rich,yuan2025self}, and PPO-based policy optimization~\cite{fan2023dpok,black2023training}. These advancements have played a pivotal role in enhancing the aesthetic quality and semantic consistency of image generation models.
Compared to image generation, RLHF research for video generation has been less explored. Some works~\cite{yuan2024instructvideo,li2024t2v} attempted to align video generation models using feedback from image reward models~\cite{wu2023human,kirstain2023pick}, but such methods struggle to capture human preference in video metrics, such as temporal consistency, motion smoothness and dynamic degree. Recent studies~\cite{wang2024lift,zhang2024onlinevpo,liu2024videodpo,xu2024visionreward,liu2025improving} aligns video generation models by training specialized video reward models on data from generated or real videos, or using existing video evaluation benchmark~\cite{huang2024vbench,he2024videoscore}. In contrast, our approach focuses on anime video generation, leveraging a dedicated anime video reward model to more effectively align generated anime content with human expectations.
\section{Method}
\label{sec:method}
\subsection{Preliminary}
\noindent{\textbf{Diffusion Models.}}
Given a sample $\mathbf{x}$ from a data distribution $p_{\text{data}}$, (\ie $\mathbf{x} \sim p_{\text{data}}$), the forward diffusion process is defined as a Markov Chain that gradually transforms $\mathbf{x}$ into gaussian noise $ \mathbf{\epsilon} \sim \mathcal{N}(0, I) $, using a predefined noise schedule $\{\alpha_t, \sigma_t\}_{t=1}^T $. This schedule governs the signal-to-noise ratio at each step.
Diffusion models are generative models that reverse this diffusion process by learning conditional distributions $p_\theta (\mathbf{x}|\mathbf{c})$, where $\mathbf{c}$ is the conditioning signal that guides the generation. The training objective of the diffusion model is to minimize the difference between the forward and reverse process:

\begin{equation}
\label{eq:diffusion}
L_{\text{DM}}(\theta) = \mathbb{E}_{x, \mathbf{\epsilon}, t} \left[ \| \mathbf{\epsilon} - \mathbf{\epsilon}_\theta(\mathbf{x}_t,\mathbf{c}) \|^2_2 \right],
\end{equation}

\noindent where $\mathbf{x}_t = \alpha_t \mathbf{x} + \sigma_t \mathbf{\epsilon}$, and $\epsilon \sim \mathcal{N}(0, I)$ is sampled gaussian noise, $t \sim \text{Uniform}\{1,..., T\}$. $\mathbf{\epsilon}_\theta$ is the denoising model parameterized by $\mathbf{\epsilon}$.

\noindent{\textbf{DPO For Diffusion Models.}}
The Bradley-Terry model is a method used to define pairwise preferences. The preference distribution $p$ can be expressed as:
\begin{equation} 
\label{eq:BT}
p(v^w \succ v^l | c)= \sigma (R( c,v^w )-R( c,v^l)),
\end{equation}
where $\sigma (\cdot)$ is the sigmoid function, $ v^w $ is the more preferred sample, $v^l$ is the less preferred sample, and $R(c,v)$ is the reward function that reflects the preference of the sample $v$ and condidtion $c$.  
DPO~\cite{rafailov2023direct} reparameterizes the reward function $R$ using a closed-form expression from the optimal policy, and expresses the probability of preference data with the policy model, yielding the following objective $\mathcal{L}_\text{DPO}(\theta)$:
\begin{footnotesize} 
\begin{equation}
\label{eq:dpo_objective}
\begin{split}
-\mathbb{E}_{(c,v^w,v^l)}\left[\log \sigma\left(\beta \log \frac{p_{\theta}\left(v^{w} \mid c\right)}{p_{\mathrm{ref}}\left(v^{w} \mid c\right)}-\beta\log \frac{p_{\theta}\left(v^{l} \mid c\right)}{p_{\mathrm{ref}}\left(v^{l} \mid c\right)}\right)\right],
\end{split}
\end{equation}
\end{footnotesize}
where $p_{\mathtt{ref}}(v|c)$ is the base reference distribution, and $\beta$ controls the distributional deviation.
However, the challenge of adapting DPO to diffusion models~\cite{ho2020denoising} is that the parameterized distribution $p(v|c)$ is not tractable. Diffusion-DPO~\cite{wallace2024diffusion} introduces approximation techniques for reverse process and propose a tractable alternative:
\begin{equation} 
\begin{split}
\label{eq:diff-dpo}
\mathcal{L}_\text{DPO}(\theta) = 
    -\mathbb{E}_{(c,v^w,v^l)} \Bigg[ 
        \log \sigma \bigg( -\beta T \omega \big(\lambda_t \big) \\ 
         \Big( 
            \big\| \epsilon^w - \epsilon_\theta (v^w_t,t) \big\|^2_2 
            - \big\| \epsilon^w - \epsilon_{\text{ref}}(v^w_t,t) \big\|^2_2 -
        \\ 
            \big( 
                \big\| \epsilon^l - \epsilon_\theta (v^l_t,t) \big\|^2_2 
                - \big\| \epsilon^l - \epsilon_{\text{ref}}(v^l_t,t) \big\|^2_2 
            \big) \Big)
        \bigg) 
    \Bigg],
\end{split}
\end{equation}
\noindent where \( v_t^* = \alpha_t v_0^* + \sigma_t \epsilon^* \), \( \epsilon^* \sim \mathcal{N}(0, I) \) is the predicted noise. \( \lambda_t = \alpha_t^2 / \sigma_t^2 \) is the signal-to-noise ratio.

\subsection{Anime Reward Datasets}
\label{sub:dataset}
The construction of traditional Text-to-Video (T2V) human preference datasets typically starts with the creation of prompt set only. However, building an Image-to-Video (I2V) human preference dataset presents a unique challenge, as it requires spatial and temporal consistency between the prompt guidance and the initial image. To address this, we propose extracting information directly from ground truth animation videos, which helps ensure precise alignment between the prompts and images. Additionally, to enhance the diversity and generalizability of the dataset, we collect animation videos spanning various action categories, such as speaking, walking, waving, kissing, crying, hugging, and pushing/pulling, \textit{etc}. 

\noindent{\textbf{Data Collection.}} We select \textbf{5k} anime video
with action labels summarized from more than 100 common actions through human annotation. Each action label corresponds to 30-50 video clips. The initial text prompts are captioned using Qwen2-VL~\cite{wang2024qwen2}, followed prompt optimization recommended in CogVideoX~\cite{yang2024cogvideox}, and the third frame of the video is extracted as the initial image.
Based on these prompts and images, we utilize \textbf{5} advanced open-source or closed-source image-to-video generation models (\ie Hailuo~\cite{minimax}, Vidu~\cite{vidu}, OpenSora~\cite{zheng2024open}, OpenSoraPlan~\cite{lin2024open}, and CogVideoX~\cite{yang2024cogvideox}) to generate anime videos. Along with \textbf{5k} Ground Truth (GT) videos, we construct a total \textbf{30k} anime video dataset for reward model training.
In addition to the training data, we construct a validation set comprising \textbf{6k} anime videos, following similar collection rules, and ensure that there is no overlap between the initial images and prompts in the validation and training set.

\noindent{\textbf{Human Annotation.}} 
\label{sec:annotation} 
To collect comprehensive human evaluations, we propose to evaluate the overall quality of generated videos from two aspect: \textit{visual appearance} and \textit{visual consistency}. Visual appearance evaluates the fundamental quality of the video, focusing on its aesthetic attributes, including visual smoothness, motion quality, and visual appeal. In contrast, visual consistency emphasizes the alignment between text and video, image and video, and the consistency of character identities.  We hired 6 annotators to annotate the dataset, rating each video on a scale from 1 to 5, with 5 representing the highest quality. The final preference score for each dimension was obtained by averaging the scores across all annotators.

\subsection{AnimeReward}
\label{sub:reward}
AnimeReward is a multi-dimensional anime reward system designed to learn human preferences of anime videos. AnimeReward consists of two aspects with six dimensions, where \textbf{visual smoothness}, \textbf{visual motion}, and \textbf{visual appeal} assess visual appearance, while \textbf{text-video consistency}, \textbf{image-video consistency}, and \textbf{character consistency} evaluate visual consistency.
Unlike approaches that rely on a single Vision-Language Model (VLM) to jointly train reward scores for all dimensions, AnimeReward employs specialized VLM for different dimensions, training them individually through reward score regression.

\subsubsection{Visual Appearance} 
\label{subsub:visual_appearance}
To provide appropriate assessment reward metrics, we utilize three observable dimensions: visual smoothness, visual motion and visual appeal.

\noindent{\textbf{Visual Smoothness.}}
Our goal is for the visual smoothness model to learn human preference for evaluating video smoothness and effectively identify distortions in animated videos.
Thus, we finetuned the vision encoder of a VLM-based~\cite{jiang2024mantis} model with a regression head to maintain consistency with manual scores. 
The formulation is shown as follows:
\begin{equation}
\label{eq:smooth}
   S_{smooth}=Reg(E_{v}(I_{1},...,I_{N})),
\end{equation}
where $I_{i}$, $N$ denote the single frame and the total frames, and $Reg$ denotes the regression head, and $E_{v}$ denotes the vision encoder.

\noindent{\textbf{Visual Motion.}} We utilize a model based on the ActionCLIP~\cite{wang2021actionclip} framework to train a motion-scoring model that evaluates the magnitude of primary motion in anime videos.
The anime video clips and their corresponding motion captions are categorized into six levels of movement amplitude, ranging from stillness to significant motion. 
When training the motion model, the designed motion prompts are listed as follows:

[\textit{'The protagonist has a large range of movement, such as running, jumping, dancing, or waving arms.',
'The protagonist remains stationary in the video with no apparent movement.'
}]

Finally, the motion score is derived from the similarity score between the designed motion prompt and the target video:
\begin{equation}
\label{eq:moiton}
    S_{motion} = Cos(MCLIP(V), MCLIP(T_{m})),
\end{equation}
where $MCLIP$ denotes the motion model. $V$ represents the generation video and $T_{m}$ denotes the designed motion prompt. 

\noindent{\textbf{Visual Appeal.}}
We define the visual appeal score to assess the fundamental quality of video generation. Previous studies utilized aesthetic score models trained on real-world datasets such as LAION-5B~\cite{schuhmann2022laion}. However, directly applying these models to anime data fails to provide sufficient differentiation, as the results across various methods appear indistinguishable.
To address this, we first extract key frames from the video using a key frame extraction method. Then, we train a regression model to learn human aesthetic standards.
The formulation shows as follows:
\begin{equation}
\label{eq:appeal}
    S_{appeal}=Aes(SigLIP(I_{0,1,\dots, K})) \quad I_{i}\in{KeyFrm(V)},
\end{equation}
where $KeyFrm$, $SigLIP$ and $Aes$ denote the key frame extraction method, feature encoder method and aesthetic evaluation method, and $K$ denotes the number of the key frames.

\subsubsection{Visual Consistency}
We consider three factors to evaluate the visual consistency of the generation video: text-video, image-video, and character consistency, respectively.

\noindent{\textbf{Text-Video Consistency.}}
To assess text-video consistency, we finetuned the vision and text encoder modules with a regression head using anime text-video pairs.
The formulation is shown as follows:
\begin{equation}
\label{eq:text-video}
    S_{tvc}=Reg(E_{v}(V), E_{t}(T)),
\end{equation}
where $Reg$ denotes the regression head, and $E_{v}$, $E_{t}$ denote the vision and text encoder, respectively.

\noindent{\textbf{Image-Video Consistency.}}
In the Image-to-Video setting, the generated video should preserve stylistic consistency with the input image.
Similar to text-video consistency evaluation, we use a vision encoder with a regression head to compute the score.
The formulation lists as follows:
\begin{equation}
\label{eq:image-video}
    S_{ivc}=Reg(E_{v}(V), E_{v}(I_{p})),
\end{equation}
where $V$ and $I_{p}$ denote the participant video clip and the input image.

\begin{figure}[!h]
    \centering
    \includegraphics[scale=0.25]{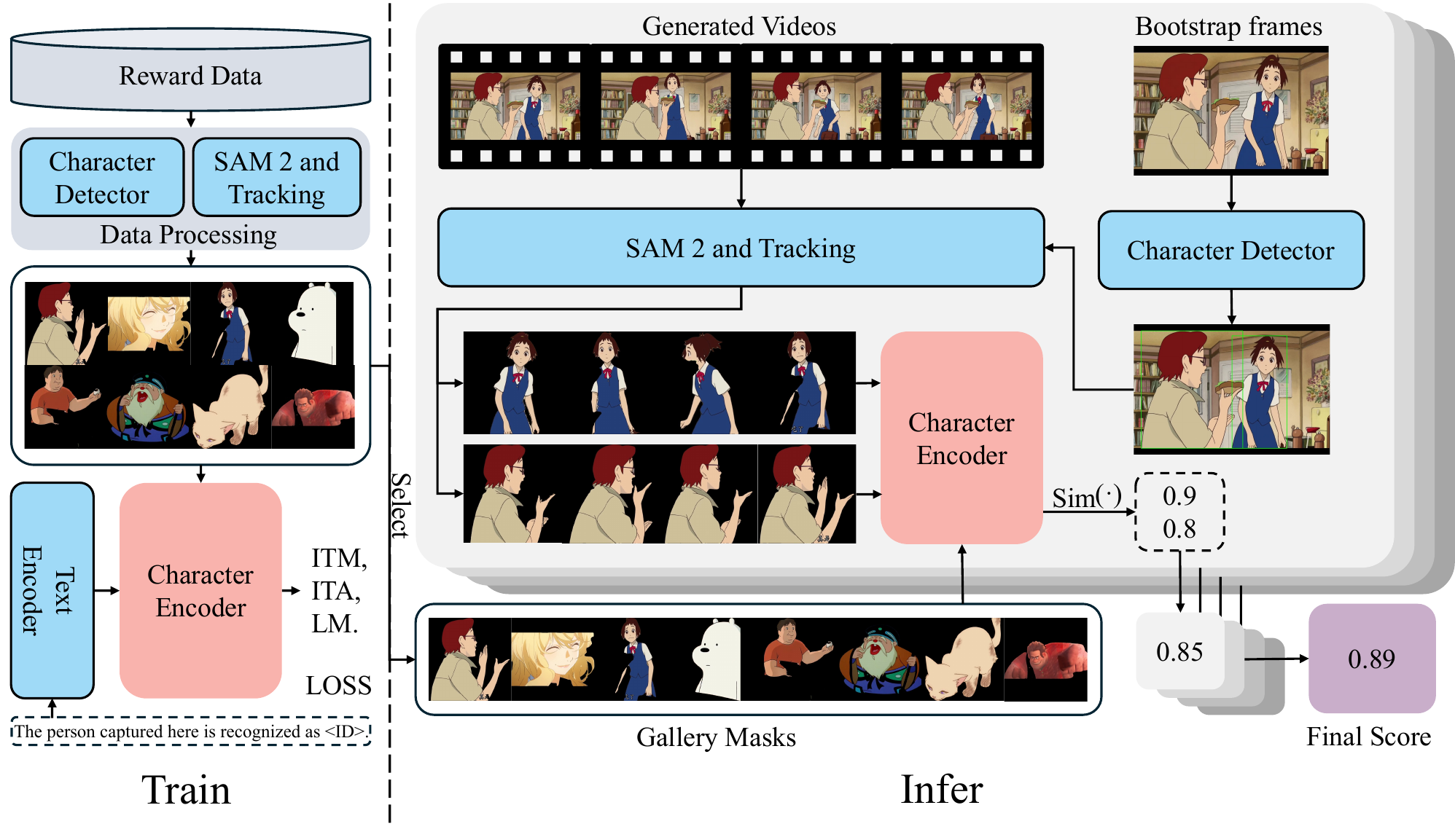}
    \caption{The framework of character consistency: training and inference.}
    \label{fig:ipconsis}
\end{figure}

\noindent{\textbf{Character Consistency.}}
Character consistency is a crucial factor in anime video generation. If the appearance of the protagonist changes throughout the animation, even a high-quality video risks potential infringement. To address this, we design a systematic procedure comprising detection, segmentation, and recognition. Fig.~\ref{fig:ipconsis} illustrates the framework of the character consistency model. We first utilize GroundingDINO~\cite{ren2024grounding}, SAM~\cite{ravi2024sam2segmentimages} and tracking tools to extract character masks for each video frame. Next, we finetune a BLIP-based model~\cite{li2022blip} to establish associations between extracted masks and their corresponding anime IP characters. 
During inference, character consistency is measured by calculating the cosine similarity between the generated character features and the stored features:
\begin{equation}
\label{eq:character}
    S_{cc}=\frac{1}{N}\sum^{N}_{i}Cos(BLIP(M_{i}), fea_{c}),
\end{equation}
where $N$ denotes the number of sample frames, $M_{i}$ denotes the mask obtained from GroudingDINO and SAM methods, and $fea_{c}$ denotes the stored character's features. 

The trained reward models across different dimensions collectively form the AnimeReward. Given a generated anime video $v$, its initial frame $x$, and the corresponding prompt $c$, the reward score for each dimension $ d \in D $ is denoted as $r_{d,\phi}(v, x, c)$, where $\phi$ is the parameter of the reward model. The final overall reward score $R(v)$ for the video $v$ is obtained by averaging the scores across all dimensions:
\begin{equation}
\label{eq:score}
    R(v) = \frac{1}{|D|} \sum_{d \in D} r_{d,\phi}(v,x,c).
\end{equation}
By learning human preferences, AnimeReward provides high-quality preference feedback for generated anime videos, serving as a guide signal for preference optimization training.

\subsection{Anime Video Alignment}
Direct Preference Optimization (DPO)~\cite{rafailov2023direct} has recently been proven effective in aligning human preferences for video generation. The training dataset in DPO consists of multiple preference sample pairs. In our I2V anime video generation alignment task, all preference samples are uniformly generated by a reference model $p_\text{ref}$, which we aim to align with.  
Given an initial image $x$ and its corresponding prompt $c$, we use the pre-trained model $p_\text{ref}$ to generate a set of $N$ videos $\{v_1,v_2,\dots,v_N\}$. Each generated video \( v_i \) is assigned a reward score $R(v_i)$ conditioned on the image $x$ and prompt $c$ using AnimeReward, following Eq.~\ref{eq:score}. The score $R(v_i)$ serves as a quantitative measure of human feedback.
When constructing preference pairs $(v^w,v^l)$, we select the video with the highest reward score as $v^w$ (the most preferred sample) and the video with the lowest score as $v^l$ (the least preferred sample) from $N$ generated videos $\{v_1,v_2,\dots,v_N\}$.

\noindent{\textbf{Gap-Aware Preference Optimization.}}
Naive DPO~\cite{rafailov2023direct} focus solely on modeling the probability $p(v^w \succ v^l | x, c)$ for each pair. However, the strategies fail to account for the magnitude of the preference gap between $v^w$ and $v^l$. To illustrate, consider two preference pairs $(v^w_1,v^l_1)$ and $(v^w_2,v^l_2)$. Assume that in AnimeReward, $v^w_1$ is scored $90$, $v^l_1$ is scored $40$, $v^w_2$ is scored $70$, and $v^l_2$ is scored $60$. In original loss function Eq.~\ref{eq:diff-dpo}, DPO treats both pairs equally, ignoring the fact that $(v^w_1,v^l_1)$ signifies a much stronger preference contrast. Ideally, pair $(v^w_1,v^l_1)$ with a greater preference gap should exert a stronger influence on the preference optimization process.

To address this issue, we propose \textbf{Gap-Aware Preference Optimization (GAPO)}, which explicitly incorporates the preference gap between winning and losing samples. 
First, we define the reward gain $G_i$ for each video:
\begin{equation}
\label{eq:gain}
    G_{i} = \alpha^{\widetilde{R}(v_i)},
\end{equation}
\noindent where $\widetilde{R}(v^i)\in [0,1]$ is the normalized reward score of generated video $v_i$ and $\alpha$ is a hyperparameter that controls the strength of the reward gain. 
For each preference pair $(v^w,v^l)$, we use the difference between the reward gain of \( v^w \) and \( v^l \) as the gap factor, which is fed back into the loss $\mathcal{L}_\text{DPO}$ in Eq.~\ref{eq:diff-dpo}, ultimately resulting in the GAPO loss:
\begin{equation}
\label{eq:gap}
    \mathcal{L}_\text{GAPO}(\theta)=(G_{w} - G_{l}) \cdot \mathcal{L}_\text{DPO}(x,c,v^w,v^l),
\end{equation}
\noindent where $G_{w}$ and $G_{l}$ are the reward gain of the winning sample \( v^w \) and the losing sample \( v^l \), respectively. By multiplying the DPO loss with the difference in reward gain between winning and losing samples, GAPO can amplify the impact of pairs with large preference gaps during training, while reducing the influence of pairs with small preference differences. This enhances the efficiency of anime video alignment, enabling the model to better capture human preference distinctions.

\begin{figure*}[t]
  \centering
  \includegraphics[width=\linewidth]{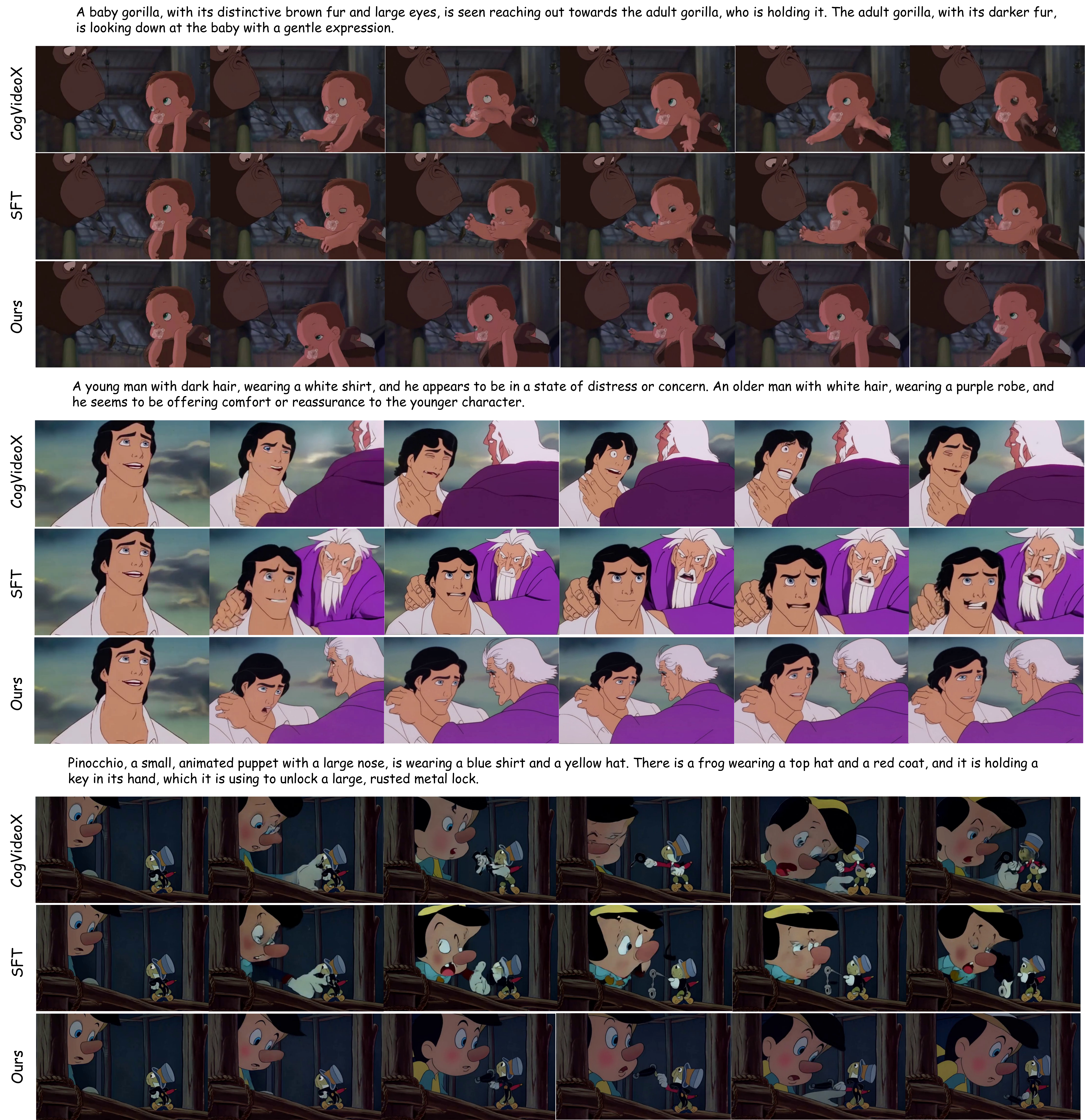}
  \caption{\textbf{Qualitative comparison results}, with the first frame in the leftmost column as the initial guiding image. Compared to the baseline and SFT models, our preference optimization method significantly reduces distortions and artifacts in anime videos while exhibiting stronger visual consistency and text alignment. The generated anime videos better align with human preferences.}
  \label{fig:quality}
\end{figure*}

\section{Experiments}
\begin{table*}[h!]
    \centering
    \begin{tabular*}{\linewidth}{@{\extracolsep{\fill}}cccccccccc@{}}
    \toprule
    \textbf{Model}  & \makecell{\textbf{Total}} & \makecell{\textbf{I2V} \\ \textbf{Subject}} & \makecell{\textbf{I2V} \\ \textbf{Back.}} & \makecell{\textbf{Subject} \\ \textbf{Consist.}} & \makecell{\textbf{Back.} \\ \textbf{Consist.}} & \makecell{\textbf{Motion} \\ \textbf{Smooth.}}
     & \makecell{\textbf{Dynamic} \\ \textbf{Degree}} & \makecell{\textbf{Aesthetic} \\ \textbf{Quality}}  & \makecell{\textbf{Imaging} \\ \textbf{Quality}} \\
    \midrule
    Baseline  & 85.73 & 95.20 & 93.15 & 93.76 & 95.49 & 98.70 & \textbf{57.33} & 53.95 & 68.44  \\
    SFT  & 85.72 & 95.53 & 93.37 & 94.12 & 95.63 & 98.87 & 47.33 & \textbf{54.55} & \textbf{69.60}  \\
    \textbf{Ours}  & \textbf{86.24} & \textbf{96.35} & \textbf{94.11} & \textbf{95.20} & \textbf{96.41} & \textbf{99.13} & 43.33 & 54.28 & 68.98  \\
    \bottomrule
    \end{tabular*}
    \vspace{-5pt}
    \caption{
        \textbf{Quantitive performance comparison on VBench-I2V~\cite{huang2024vbench}}, with best performance in {\bf bold}. Our alignment method achieves the highest total score, with significant improvements in visual consistency and motion stability. Without relying on any external videos for training, our method outperforms the baseline model in almost all evaluation metrics.
    }
    \label{tab:vbench}
\end{table*}

\begin{table*}[!h]
    \centering
    \setlength{\tabcolsep}{0pt}
    \begin{tabular*}{\linewidth}{@{\extracolsep{\fill}}cccccccccccc@{}}
    \toprule
    \multirow{3}{*}{\textbf{Method}} & \multicolumn{6}{c}{\textbf{AnimeReward}} &\multicolumn{5}{c}{\textbf{VideoScore}}\\
    \cmidrule(r){2-7} \cmidrule(r){8-12}& \makecell{\textbf{Visual}\\ \textbf{Smooth}} & \makecell{\textbf{Visual}  \\ \textbf{Motion}} & \makecell{\textbf{Visual} \\ \textbf{Appeal}} & \makecell{\textbf{Text} \\ \textbf{Consist.}} & \makecell{\textbf{Image} \\ \textbf{Consist.}} & \makecell{\textbf{Charac.} \\ \textbf{Consist.}} & \makecell{\textbf{Visual} \\ \textbf{Quality}} & \makecell{\textbf{Temporal} \\ \textbf{Consist.}}& \makecell{\textbf{Dynamic} \\ \textbf{Degree}} & \makecell{\textbf{Text} \\ \textbf{Alignment}}& \makecell{\textbf{Factual} \\ \textbf{Consist.}}\\
    \midrule
    Baseline  & 52.63 & 65.13 & 47.68 & 71.50 & 74.92 & 88.85 &3.361 &3.080 & \textbf{3.334} &2.945 &  3.114 \\
    SFT  & 52.62 & \textbf{66.27} & 48.11 & 72.81 & 75.25 & 90.38  &  3.358 & 3.075 &3.329 &\textbf{2.951} & 3.100 \\
    \textbf{Ours}  & \textbf{61.67} & 59.86 & \textbf{53.08} & \textbf{73.29} & \textbf{79.30} & \textbf{91.45} &\textbf{3.379} &\textbf{3.131} & 3.290 &2.942 & \textbf{3.148} \\
    \bottomrule
    \end{tabular*}
    \vspace{-5pt}
    \caption{
        {\textbf{Quantitive performance comparison on AnimeReward and VideoScore}~\cite{he2024videoscore}}, with best performance in {\bf bold}. We achieve the best performance in both visual quality and motion smoothness, with a higher degree of distinction on AnimeReward.}
    \label{tab:animereward}
\end{table*}

\subsection{Experiment Setup}
\noindent{\textbf{Dataset.}}
We utilize the open-source CogVideoX-5B~\cite{yang2024cogvideox} model as the baseline for our alignment experiments. 
Following the data collection strategy outlined in Sec.~\ref{sub:dataset}, we construct an initial training set consisting of \textbf{2k} original anime images with their corresponding prompts. Based on these data, we employ the baseline model to generate $N=4$ anime videos for each data instance. To simulate human feedback, we leverage our AnimeReward to assess all generated videos and assign reward scores. Within each set of four videos, we select the highest-scoring and lowest-scoring videos to form a preference pair, ultimately creating a training data with \textbf{2k} preference sample pairs.
Noting that all training videos are generated exclusively by the baseline model without any real video data. This further underscores the effectiveness of our alignment pipeline. 

\noindent{\textbf{Training Settings.}}
In alignment experiments, we finetune all transformer blocks to better align the model with human preferences. The global batch size is set to $8$, and  that of learning rate is $5e-6$. The DPO hyperparameter $\beta$ is $5000$, and the GAPO hyperparameter $\alpha$ is $2$. 
For supervised fine-tuning (SFT), we finetune the baseline model using the highest reward-scoring video. All experiments are conducted on $8$ A800 GPUs. The generated videos are $49$ frames, $16$ fps, with a resolution of $480\times720$.

\noindent{\textbf{Evaluation Metrics.}}
Both automated metrics and human evaluation are applied to assess model capabilities. The automated evaluation consists of three methods: VBench-I2V~\cite{huang2024vbench}, VideoScore~\cite{he2024videoscore}, and our proposed AnimeReward. 
For human evaluation, each sample is assessed by three annotators. Given a pair of videos $\{v_a,v_b\}$, the annotators can choose from three possible judgments: (1) $v_a$ is better than $v_b$, (2) $v_a$ is worse than $v_b$, or (3) $v_a$ and $v_b$ are tied. Video $v_a$ is considered to win or lose against $v_b$ only if at least two out of the three annotators agree that $v_a$ is better or worse than $v_b$, respectively. In all other cases, the pair $\{v_a,v_b\}$ is considered a tie. 
\begin{figure}[!t]
  \centering
  \includegraphics[width=0.9\linewidth]{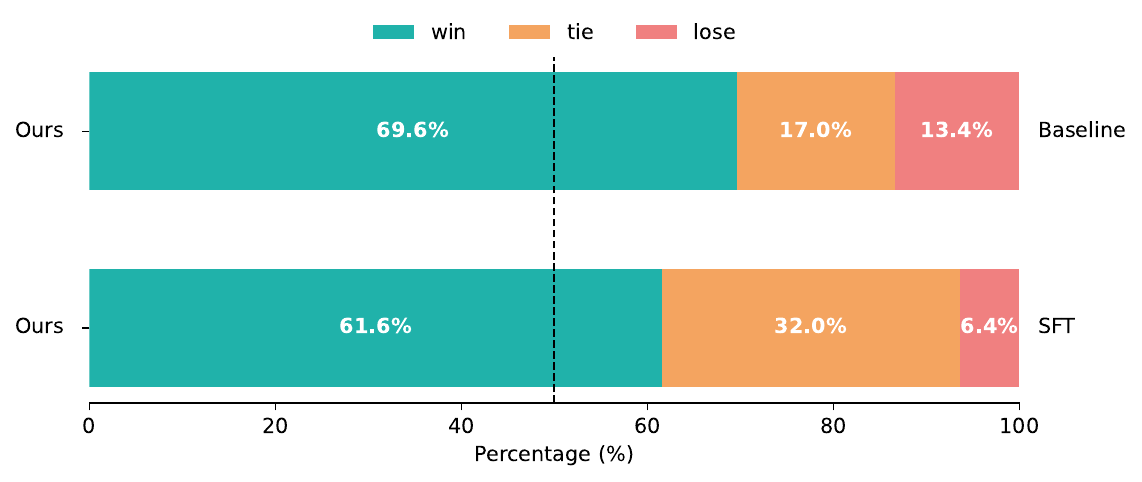}
  \vspace{-5pt}
  \caption{Human evaluation of generated anime videos on different models.}
  \label{fig:quanti_human}
\end{figure}

We follow the data collection method in Sec.~\ref{sub:dataset}, gathering $500$ images along with corresponding prompts as the test set. We then generate anime videos using the baseline model, the SFT model, and our GAPO model, allowing for a comparative analysis of their performance. 

\subsection{Experiment Result}
\noindent{\textbf{Quantitative Results.}}
Tab.~\ref{tab:vbench} and Tab.~\ref{tab:animereward} present the quantitative comparison results on VBench-I2V~\cite{huang2024vbench}, AnimeReward, and VideoScore~\cite{he2024videoscore}. 
On VBench-I2V, our preference alignment method achieves the highest total score, consistently outperforming the baseline across nearly all metrics and surpassing the SFT model in most cases. Notably, it demonstrates significant improvements in ``I2V Subject" and ``Subject Consistency", indicating a stronger ability to maintain anime character consistency, which is a crucial factor in I2V anime video generation.
In the AnimeReward evaluation, our method exhibits significant improvements across all metrics except visual motion, indicating that it better aligns with human preferences in terms of visual appearance and consistency. 
For VideoScore, our alignment strategy outperforms both the baseline and SFT models in three dimentions, demonstrating superior temporal stability and content fidelity. 
Additionally, we observe that our model performs worse than the baseline and SFT methods in ``Dynamic Degree" (\ie ``\textit{Visual Motion}"). 
We argue that videos with higher dynamic degrees are more prone to distortions and artifacts, which in turn degrade overall visual quality and negatively impact human preference ratings.  
These results indicate that humans generally tend to favor videos with superior visual quality, stronger consistency, and greater stability.
Fig.~\ref{fig:quanti_human} presents the human evaluation results on \textbf{500} anime videos generated by different models. In the human preference comparisons, our alignment model achieves a significantly higher win rate than both the Baseline and SFT models, exceeding 60\%. Although the SFT model is trained on the highest-scoring winning samples, its generated video quality does not improve; instead, its win rate is even lower than that of the Baseline model. This further demonstrates the superior performance of our GAPO method in aligning with human preferences.

\noindent{\textbf{Qualitative Results.}}
Fig.~\ref{fig:quality} presents the visual comparison results. It can be observed that after applying preference alignment finetuning to the baseline model, our method significantly reduces distortions and artifacts in the generated videos, achieving higher motion stability and character consistency, and resulting in substantially better visual quality compared to both the baseline and SFT methods.
Although the baseline and SFT models exhibit higher motion dynamics, such as the baby raising both hands in the first case and Pinocchio repeatedly extending his hand in the third case, the larger motion amplitudes also introduce a significant number of face distortions and motion artifacts, which severely impact the visual appeal.
Additionally, our method demonstrates superior text consistency compared to other methods. For instance, in the second case, the prompt specifies that an older man in a purple coat is comforting a young man. However, in the results from the baseline and SFT methods, the older man appears to be threatening or intimidating the young man, who shows a frightened expression. In contrast, our method correctly generates the older man patting the young man's shoulder in a comforting manner.
From a visual perspective, the anime videos generated by our method align more closely with human preferences, further underscoring the effectiveness of our approach in human preference alignment.

\subsection{Ablation Studies}
\noindent{\textbf{Effect of Gap-Aware Preference Optimization.}}
To verify the advantages of Gap-Aware Preference Optimization (GAPO) versus Direct Preference Optimization (DPO)~\cite{rafailov2023direct}, we conduct comparative experiments using the same experimental setup but different preference optimization algorithm. We evaluated the models on above three evaluation methods. The results are shown in Tab. ~\ref{tab:abla_gapo}, where AR and VS adopt average scores across dimensions as the final evaluation result. 
GAPO achieves the best performance across all three evaluations, with significant improvements over the DPO in V-I2V and AR. These results demonstrates that GAPO can improve preference alignment efficiency and guide the model to generate anime videos that better align with human preferences.

\begin{table}[!h]
\centering
  \begin{tabular*}{0.85\columnwidth}{@{\extracolsep{\fill}}cccc}
    \toprule
    \textbf{Method} &  V-I2V  & AR  & VS \\
    \midrule
    Baseline & 85.73 & 66.78 &  3.167 \\
    DPO & 86.07 & 67.81 & 3.176 \\
    GAPO & \textbf{86.24} & \textbf{69.77} & \textbf{3.178} \\
    \bottomrule
  \end{tabular*}
  \vspace{-5pt}
\caption{Ablation study results of \textbf{GAPO} on the three evaluation criteria. V-I2V refers to VBench-I2V~\cite{huang2024vbench}, AR stands for AnimeReward, and VS represents VideoScore~\cite{he2024videoscore}.\textbf{DPO} refers to aligning the baseline model using the naive DPO algorithm.}
\label{tab:abla_gapo}
\end{table}

\begin{figure}[!h]
    \centering
    \includegraphics[width=0.9\linewidth]{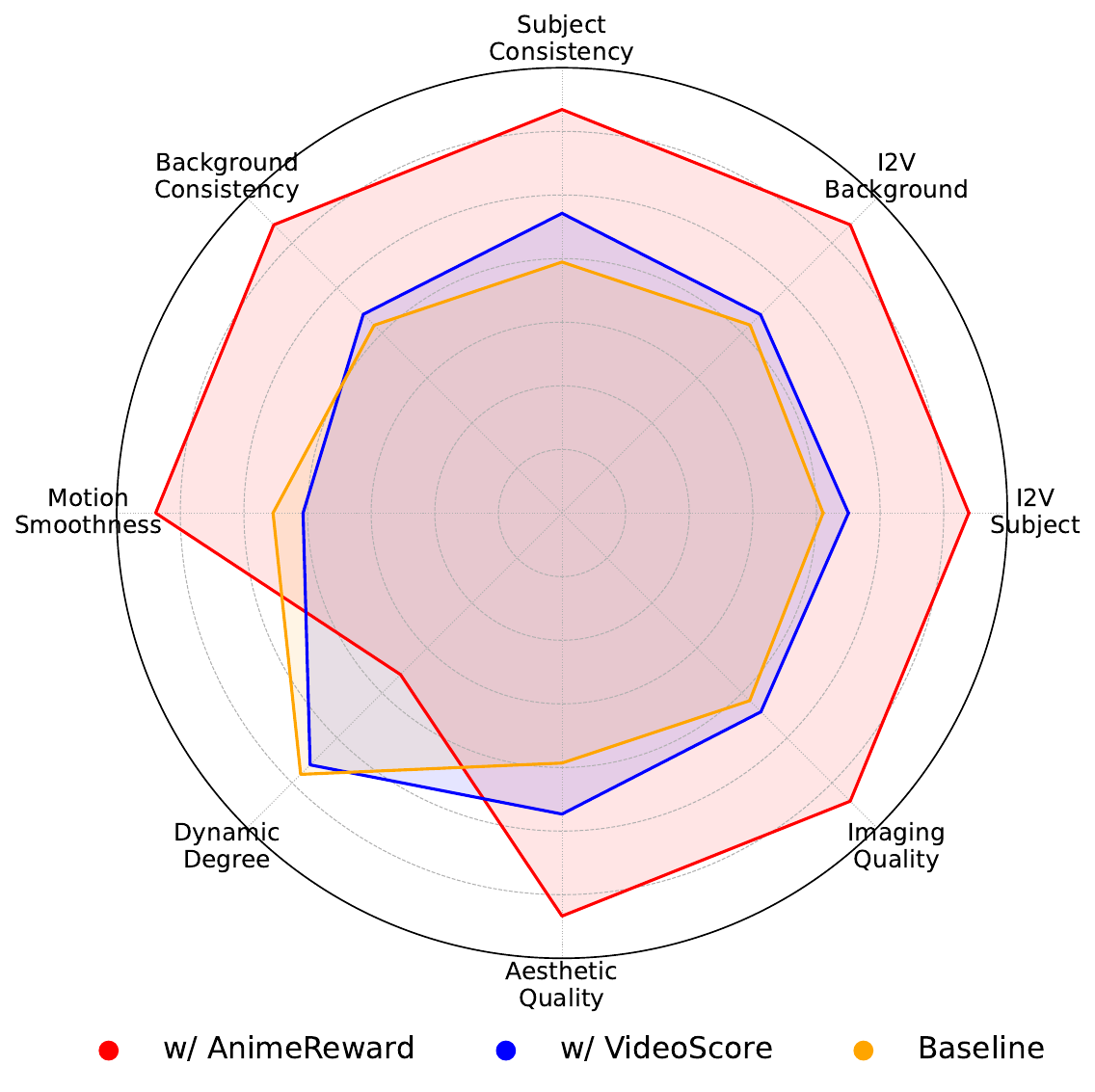}
    \caption{Visualized evaluation results on multiple dimensions of VBench-I2V~\cite{huang2024vbench} based on different reward models.}
    \label{fig:radar_reward}
\end{figure}

\begin{table}[h]
\centering
  \begin{tabular*}{0.9\columnwidth}{@{\extracolsep{\fill}}lccc}
    \toprule
    \textbf{Reward} &  V-I2V  & AR  & VS \\
    \midrule
    VideoScore & 85.81 & 67.65 & \textbf{3.181} \\
    AnimeReward & \textbf{86.24} & \textbf{69.77} & 3.178\\
    \bottomrule
  \end{tabular*}
  \vspace{-5pt}
\caption{Ablation study results based on different reward models.}
\label{tab:abla_reward}
\end{table}

\begin{figure}[!h]
    \centering
    \includegraphics[width=\linewidth]{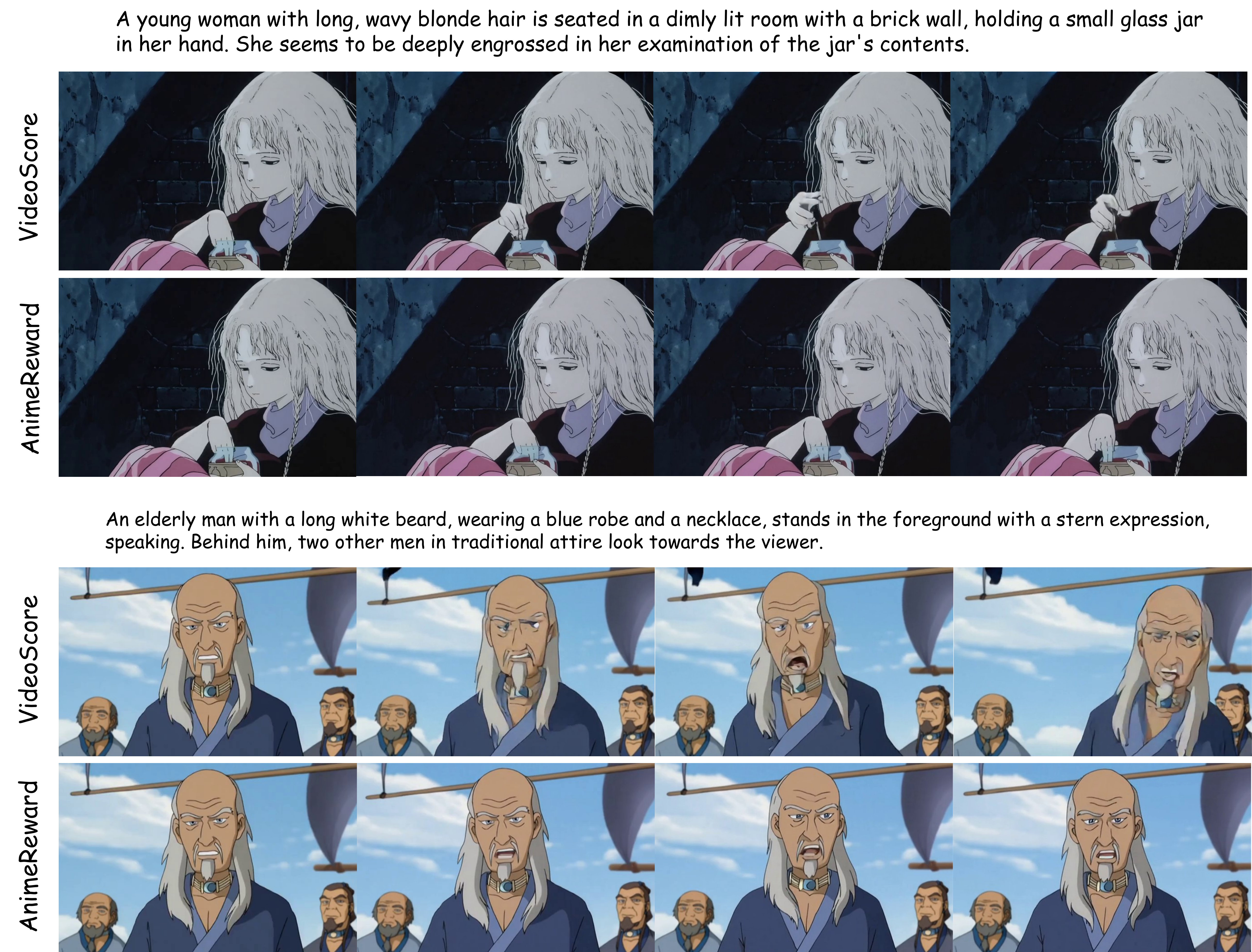}
    \caption{Visual comparison of alignment generation results based on AnimeReward and VideoScore~\cite{he2024videoscore} reward models.}
    \label{fig:able_videoscore}
\end{figure}

\noindent{\textbf{Comparison of Reward Models.}}
To verify the superiority of AnimeReward in preference alignment, we conducted an experiment using VideoScore as the reward model. The results are presented in Tab.~\ref{tab:abla_reward}.
The model trained with AnimeReward outperforms VideoScore on two evaluation criteria, whereas VideoScore performs better primarily within its own evaluation framework. For a more objective assessment, we show the multidimensional results on the third-party evaluation VBench-I2V in Fig.~\ref{fig:radar_reward}. The alignment model trained with AnimeReward outperforms VideoScore across seven dimensions except dynamic degree. This further validates that AnimeReward can provide more accurate human preference feedback for anime videos. Additionally, we present a visual comparison of both reward methods in Fig.~\ref{fig:able_videoscore}. While the results based on VideoScore exhibit greater motion dynamics, such as the girl's hand extending beyond the cup and the elderly man's head shaking more significantly, these movements do not align with the given text guidance. Instead, they introduce noticeable distortions and artifacts in the generated visual content. In contrast, our AnimeReward-based results achieve stronger text-video consistency and greater visual appeal.

\begin{table}[t]
\centering
  \begin{tabular*}{0.9\columnwidth}{@{\extracolsep{\fill}}lccc}
    \toprule
    \textbf{Proportion} &  V-I2V  & AR  & VS \\
    \midrule
    0:0:1:1:1:1 & 85.90 & 68.17 & 3.174 \\
    3:5:3:3:3:3 & 85.81 & 67.71 & \textbf{3.188}\\
    3:3:6:4:4:4 & 86.14 & 68.59 & 3.176 \\
    1:1:1:1:1:1 & \textbf{86.24} & \textbf{69.77} & 3.178 \\
    \bottomrule
  \end{tabular*}
  \vspace{-5pt}
\caption{Ablation study results on different weighting strategies for reward scores.}
\label{tab:abla_weight}
\end{table}

\noindent{\textbf{Different Weighting Strategies for Reward Scores.}}
To validate the correctness of using the average score across all dimensions as the final reward score in Eq.~\ref{eq:score}, we explore the alignment effectiveness under different weighting strategies for the six dimensions in AnimeReward. The comparison results are presented in Tab.~\ref{tab:abla_weight}. In addition to the average weighted strategy (the fourth row), we evaluate the following alternative weighting schemes:
(1) Ignore dynamic dimensions (the first row), where smoothness and motion are excluded from the reward weighting.
(2) Increase the weight of motion score (the second row), where the contribution of dynamic dimension (\ie \textit{visual motion}) to the final reward is increased.
(3) Prioritise non-motion dimensions with an emphasis on \textit{visual appeal} (the third row), where dimensions other than motion-related receive higher weighting, with a particular focus on \textit{visual appeal}.

\begin{figure}[t]
    \centering
    \includegraphics[width=\linewidth]{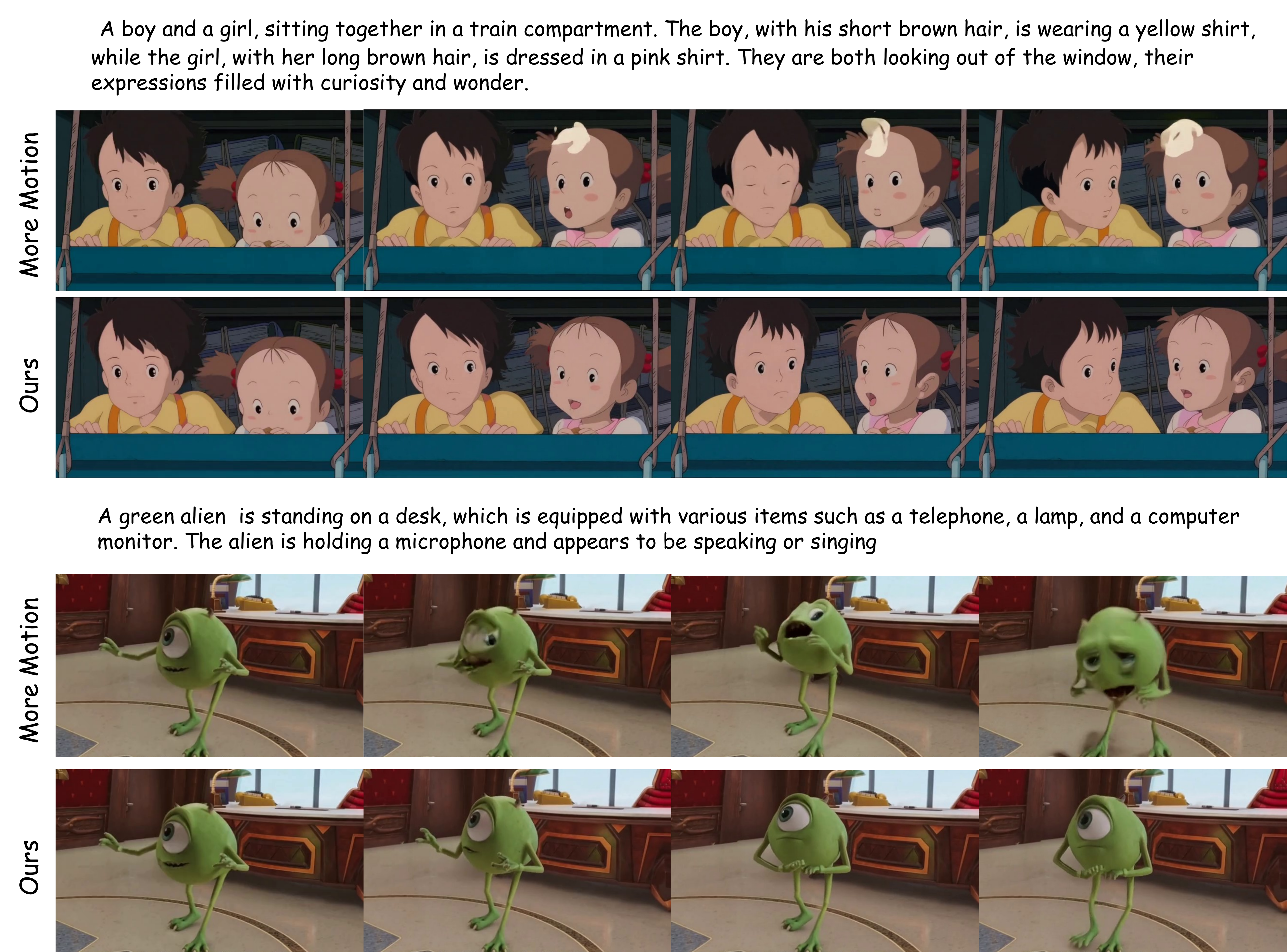}
    \caption{Visual comparison results between the average weighted strategy (Ours) and the strategy with increased motion score weight (More motion).}
    \label{fig:able_motion}
\end{figure}

\begin{table*}[h]
    \centering
    \begin{tabular*}{\linewidth}{@{\extracolsep{\fill}}cccccccccc@{}}
    \toprule
    \textbf{Strategy}  & \makecell{\textbf{Total}} & \makecell{\textbf{I2V} \\ \textbf{Subject}} & \makecell{\textbf{I2V} \\ \textbf{Back.}} & \makecell{\textbf{Subject} \\ \textbf{Consist.}} & \makecell{\textbf{Back.} \\ \textbf{Consist.}} & \makecell{\textbf{Motion} \\ \textbf{Smooth.}}
     & \makecell{\textbf{Dynamic} \\ \textbf{Degree}} & \makecell{\textbf{Aesthetic} \\ \textbf{Quality}}  & \makecell{\textbf{Imaging} \\ \textbf{Quality}} \\
    \midrule
     More Motion  & 85.81 & 95.43 & 93.29 & 94.16 & 95.54 & 98.58 & \textbf{56.00} & 54.01 & 68.45  \\
    \textbf{Ours}  & \textbf{86.24} & \textbf{96.35} & \textbf{94.11} & \textbf{95.20} & \textbf{96.41} & \textbf{99.13} & 43.33 & \textbf{54.28 }& \textbf{68.98}  \\
    \bottomrule
    \end{tabular*}
    \caption{
        {Quantitative evaluation results} of the average weighted strategy (Ours) and the strategy that increases the weight of motion score (More motion) on VBench-I2V~\cite{huang2024vbench}.
    }
    \label{tab:vbench_moremotion}
\end{table*}

\begin{table*}[h]
    \centering
    \setlength{\tabcolsep}{0pt}
    \begin{tabular*}{\linewidth}{@{\extracolsep{\fill}}cccccccccccc@{}}
    \toprule
    \multirow{3}{*}{\textbf{Method}} & \multicolumn{6}{c}{\textbf{AnimeReward}} &\multicolumn{5}{c}{\textbf{VideoScore}}\\
    \cmidrule(r){2-7} \cmidrule(r){8-12}& \makecell{\textbf{Visual}\\ \textbf{Smooth}} & \makecell{\textbf{Visual}  \\ \textbf{Motion}} & \makecell{\textbf{Visual} \\ \textbf{Appeal}} & \makecell{\textbf{Text} \\ \textbf{Consist.}} & \makecell{\textbf{Image} \\ \textbf{Consist.}} & \makecell{\textbf{Charac.} \\ \textbf{Consist.}} & \makecell{\textbf{Visual} \\ \textbf{Quality}} & \makecell{\textbf{Temporal} \\ \textbf{Consist.}}& \makecell{\textbf{Dynamic} \\ \textbf{Degree}} & \makecell{\textbf{Text} \\ \textbf{Alignment}}& \makecell{\textbf{Factual} \\ \textbf{Consist.}}\\
    \midrule
    More Motion & 55.56 & \textbf{65.92} & 49.36 & 70.54 & 74.70 & 90.21  &  \textbf{3.383} & 3.107 &\textbf{3.350} &\textbf{2.963} & 3.137 \\
    \textbf{Ours}  & \textbf{61.67} & 59.86 & \textbf{53.08} & \textbf{73.29} & \textbf{79.30} & \textbf{91.45} &3.379 &\textbf{3.131} & 3.290 &2.942 & \textbf{3.148} \\
    \bottomrule
    \end{tabular*}
    \caption{
        {Quantitative evaluation results} of the average weighted strategy (Ours) and the strategy that increases the weight of motion score (More motion) on AnimeReward and VideoScore~\cite{he2024videoscore}.
    }
    \label{tab:animereward_moremotion}
\end{table*}

The results in Tab.~\ref{tab:abla_weight} indicate that the average weighted strategy achieves the highest scores on both V-I2V and AR. In contrast, increasing the motion score weight yields the best performance on VS, suggesting that this strategy aligns better with the preferences of the VS.
To further analyze the strategy of increasing motion weight, we provide detailed performance across individual dimensions on all three evaluations metrics in Tab.~\ref{tab:vbench_moremotion} and Tab.~\ref{tab:animereward_moremotion}. We find that the strategy of more motion weight compared to our average weighted strategy indeed achieves the highest scores in all dynamics-related (visual motion in AnimeReward) dimensions across different evaluation criteria. However, it underperforms our average weighted strategy across all other dimensions in VBench-I2V and AnimeReward. Fig.~\ref{fig:able_motion} additionally presents the visual comparison results. In the first case, the more motion strategy generates a strange white circle on the girl's forehead.
In the second case, the big-eyed character’s face exhibits severe distortion while speaking. In contrast, our average weighted strategy successfully completes the corresponding actions while maintaining visual aesthetics and smooth motion, which suggests that higher motion does not always contribute positively to the overall video quality. 

\section{Conclusion}
\label{sec:conclusion}
In this paper, we propose an AnimeReward function to give a comprehensive assessment of anime video generation.
Six perceptual dimensions are carefully designed to evaluate the overall quality of the generated animation videos.
With the help of AnimeReward, we build a novel training technique, Gap-Aware Preference Optimization (GAPO), which explicitly incorporates preference gaps into the optimization process to further improve the alignment performance. Experiments demonstrate that utilizing only data generated by the baseline model, our alignment pipeline significantly enhances the quality of anime generation, achieving better alignment with human preferences. 

{   
    \small
    \bibliographystyle{ieeenat_fullname}
    \bibliography{main}
}
\end{document}